\documentclass[twocolumn]{article}
\usepackage{graphicx} 
\usepackage{placeins}
\usepackage{subcaption}
\usepackage[margin=1in]{geometry}
\usepackage{booktabs}
\usepackage{amsmath}
\usepackage{amssymb}
\usepackage{hyperref}
\usepackage{xurl}
\usepackage{authblk}
\usepackage{microtype}
\usepackage{tabularx}
\usepackage{array}

\usepackage{xcolor}
\usepackage{balance}

\newcolumntype{L}{>{\raggedright\arraybackslash}X}

\title{NoduLoCC2026: Lung Nodule Localization and Classification Contest from Chest X-Ray Images}
\author[1]{Adnan Mustafic}
\author[2]{Halim Benhabiles}
\author[3]{Adnane Cabani}
\author[4]{Kristhian André Oliveira Aguilar}
\author[5]{Romain Amigon}
\author[5]{Clément Bardin}
\author[6]{Chiara Bentifece}
\author[5]{Marin Boehm}
\author[5]{Kévin Bouchard}
\author[6]{Laura Burattini}
\author[4]{Diedre Carmo}
\author[7]{Fahima Idiri}
\author[5]{Matthis Lahargoue}
\author[6]{Ilaria Marcantoni}
\author[8]{Hicham Messaoudi}
\author[1]{Cyril Meyer}
\author[9]{Farid Meziane}
\author[5]{Léon Morales}
\author[4]{Letícia Rittner}
\author[6]{Agnese Sbrollini}
\author[5]{Léonard Zipper}
\author[1]{Karim Hammoudi}
\affil[1]{Université de Haute-Alsace, IRIMAS, 68100 Mulhouse, France}
\affil[2]{IMT Nord Europe, Institut Mines-Télécom, Université de Lille, Center for Digital Systems, F-59000 Lille, France}
\affil[3]{Université de Rouen Normandie, ESIGELEC, IRSEEM, 76000 Rouen, France}
\affil[4]{Universidade Estadual de Campinas, Campinas, Brazil}
\affil[5]{Université du Québec à Chicoutimi, Chicoutimi, Canada}
\affil[6]{Università Politecnica delle Marche, Ancona, Italy}
\affil[7]{Université de Bejaia, Faculté des Sciences Exactes, Laboratoire d’Informatique Médicale et des Environnements Dynamiques et Intelligents (LIMED), Bejaia 06000, Algeria}
\affil[8]{Université de Bretagne Occidentale, Brest, France}
\affil[9]{University of Derby, Derby, United Kingdom}

\newcommand{\B}[1]{\textbf{#1}}

\date{}
\begin{document}

\maketitle

\begin{abstract}
We propose NoduLoCC2026, a challenge on lung nodule detection and localization in chest X-ray images.
We have provided a dataset for both tasks and received submissions from 5 international teams.
The participating teams' solutions are presented in this work along with results on an external dataset used for testing. 
Proposed methods show good performance on the classification task. 
The best method shows a balanced accuracy score of 0.72 and AUC-ROC of 0.79. We highlight the limitations of current approaches for the localization task, with the best approach having predicted the correct number of nodules on 53\% of the test images with a median distance of 12.83mm, showing that it is a more challenging task than the first one.
The challenge website is available via \url{https://gt-i2mdp.github.io/website/nodule_challenge.html}.

\end{abstract}

\section{Introduction}
Lung nodules are growths within the lungs that show up as opacities in Chest X-Ray (CXR) scans. These can be benign or cancerous, meaning a timely and efficient diagnosis is critical \cite{macmahon2017guidelines, national2011reduced}.

According to the Global Cancer Observatory, lung cancer remains the leading cause of both cancer incidence and mortality worldwide, with approximately 2.4 million new cases and 1.8 million deaths reported in 2022 alone \cite{who_cancer_deaths}. 
Furthermore, recent data from the Institute for Health Metrics and Evaluation, based on the Global Burden of Disease study, indicate that lung cancer accounts for 578 disability-adjusted life years (DALYs) lost per 100,000 people—higher than any other cancer type \cite{ihme_gbd_cancer_2025}.


Furthermore, Computed Tomography (CT) scans are more expensive, time consuming and expose the patient to more radiation than CXR scans \cite{brenner2007computed, nam2019development}, leading to the question of whether it is possible to automatically detect and localize lung nodules within CXR as well as we can in CT scan data.

CXR analysis for nodule detection is inherently more difficult than in CT scans due to the 3D to 2D projection nature of the image showing  overlapping structures \cite{blind_spots}. 
This makes diagnosis difficult in some cases as overlapping regions of similar densities are difficult to differentiate \cite{emphysema, design_cxr}. 

These difficulties did not slow down the advances in automated CXR analysis.
Explorations using deep learning for CXR analysis started with convolutional neural networks (CNN) \cite{chexnet, cxr8}.
Although powerful, CNNs often lack a global view of the image leading to analyses based on aggregation of small features \cite{image_is_worth_16, replace_cnn}.
The limits of such approaches were then surpassed using the vision transformer (ViT) paradigm, which allows the model to generate a more global view of the input image \cite{dinocxr}.
More recently, Mamba-based architectures started to challenge transformers and convolutions within natural and biomedical imaging using a more easily scalable paradigm than transformers \cite{mamba, vmamba, umamba}.
The most recent approaches now focus on generating a complete radiological report based on the input CXR image.
These types of models, known as vision-language models (VLM), are often very large compared to non generative propositions \cite{raddino2025, sellergren2025medgemma, chexagent-2024}. 
These most recent developments on VLMs are only made possible due to massive datasets like Chest X-ray 14, MIMIC-CXR, CheXpert, and PadChest \cite{cxr8, mimic-cxr-paper, chexpert, padchest}.

These datasets unfortunately have their own internal biases inherent to when, where, and how they were gathered \cite{bias_1, scanner_bias}. 
No single dataset can by itself guarantee a perfect distribution of patients showing a need for a sequestered test set to show true generalization capabilities to other patient demographics and scanner types \cite{Kelly2019-gr, pooch2020}.

We proposed NoduLoCC2026, an international competition for lung nodule classification and localization in chest X-ray images. 
We have restrained the participating teams to be composed of at least one expert, as stated on the challenge website, the competition was open to ``any teams from universities, research institutes, or industry labs of which at least one member holds a Ph.D in computer science, biomedical engineering, or radiology''.
This ensured high quality submissions.
Out of 20 participating teams, only 5 provided an approach to solve the classification task, and out of these teams, only four proposed a solution for the localization task. 
This already indicates the high degree of difficulty of our proposed challenge.

We proposed two tasks for this challenge.
The first task is nodule detection modeled as a binary classification task where the objective is to find whether a nodule is present or absent in a certain CXR. 
Performance of this task is measured via balanced accuracy, AUC-ROC, recall, specificity and precision.
The second task is localization. 
It has the objective of localizing nodules as precisely as possible in a CXR image. 
The performance of this task is measured using coverage, which we define as the proportion of images where the number of predictions matches the number of ground truths present in the image, as well as using euclidean distance measured in mm. 

\subsection{Datasets}
\subsubsection{Provided Train Set}
\begin{figure}
    \centering
    \includegraphics[width=\linewidth]{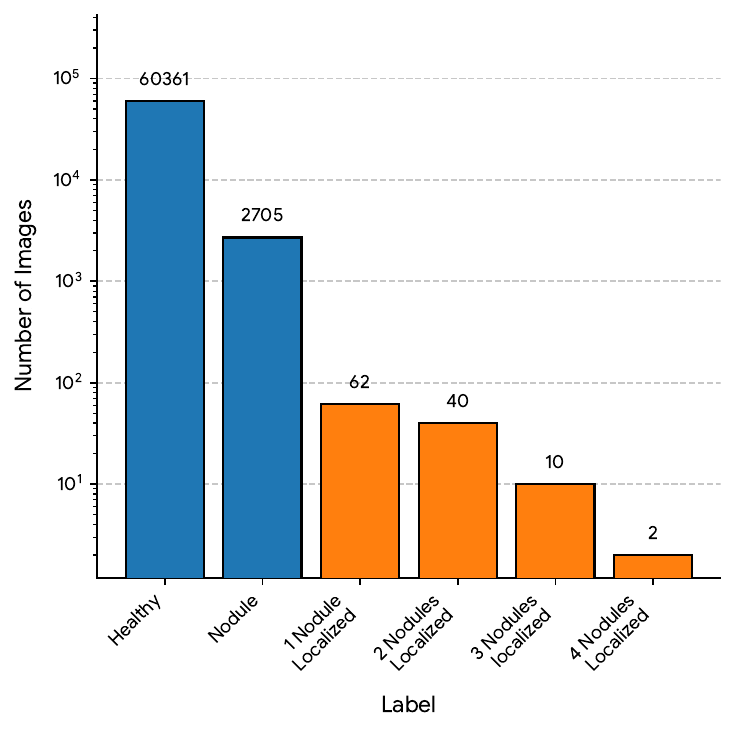}
    \caption{Log scale data distribution chart for the provided training set. Blue represents image-wise annotations, orange represents localization annotation.}
    \label{fig:data_distribution}
\end{figure}
As training sets, we have provided images from two distinct sources. 
The first source is from the NIH ChestX-ray14 dataset \cite{cxr8}, the second is from the LIDC-IDRI dataset \cite{lidc-idri}.

As the NIH ChestX-ray14 dataset is inherently multi-class and multi-label, we filtered the images to include only those that are either strictly normal or explicitly labeled as containing nodules without any additional conditions. 
This was done to ensure that other pathologies do not interfere with diagnostic performance due to visual similarities or potential occlusions. 
The images are provided as \mbox{8-bit} PNG files with a resolution of $1024 \times 1024$ pixels, representing the highest quality available from the source.


The LIDC-IDRI dataset is usually used as a CT scan database due to the existing segmentation annotations of nodules between 3mm and 30mm in diameter.
These sizes were chosen as any nodular growth over 30mm in diameter is considered a mass and any under 3mm is considered a micro-nodule which is harmless in most cases \cite{lidc-idri}.
This dataset also contains CXR images for a subset of these images, 290 CXR for 1018 CT scans. 
Nodule center point annotations are available for a subset of these images. 
We were able to identify 113 images for which the number of annotated nodules is the same as the number of segmented nodules in the corresponding CT scan.
Each of the center points was annotated four times and we have provided the mean point of these annotations for each nodule.

These images were provided as non normalized \mbox{16-bit} png files with varying resolutions, most of them hover at $2022\times2022$ pixels. 
This was done to provide a single file type for both tasks while preserving image quality, represented by bit-depth and resolution. 
No normalization was applied to avoid denaturing the data.
This meant using a bit depth of 14 within a file type supporting \mbox{16-bit} depth.
This makes the images appear dark to the eye.
The data distribution of the provided dataset is shown in Figure \ref{fig:data_distribution}.

\subsubsection{Test Set}
As a test set, we use the JSRT dataset \cite{jsrt}. This dataset is composed of 154 nodule-positive and 93 nodule-negative images. This set is used in full for the classification task and only the nodule-positive images are used for the localization task. 
The \mbox{12-bit} $2048\times2048$ pixel images were saved as \mbox{16-bit} png files.
This was done to minimize pre-processing differences with the provided LIDC-IDRI images. 
The test images were then passed through the inference scripts proposed by the participating teams.

\section[Participating Teams and Methods]{Participating Teams and \protect\\ Methods}

\begin{table*}
    \centering
    \begin{tabularx}{\textwidth}{l LLLL}
        \toprule
         Team & Backbone Architecture & Loss Function & Class Imbalance Strategy & Ensemble \& Inference Strategy \\
         \midrule
         CM@MSD & EfficientNet (B5, B6, B7) & BCE & Oversampling (1:1 ratio) & 4-model ensemble (average), threshold: 0.25 \\
         LAiB & ConvNeXtV2-B & Weighted BCE ($w_{pos}=2.0$) & Weighted sampling (1:1 ratio) \& Mixup & 5-fold ensemble, TTA (20 passes) \\
         LIMED & RadDINO ViT & Asymmetric Sigmoid & Balanced sampling (1:1 ratio) & 5-fold checkpoint merge (greedy soup) \\
         MICLab & MedGemma 1.5 (4B VLM) & Weighted CE ($w_{pos}=20$) & Balanced sampling (1:1 ratio) & Single model (QLoRA 4-bit) \\
         UQAC & EfficientNetV2-S & Focal Loss & Weighted sampling & Single model, tuned threshold \\
         \bottomrule
    \end{tabularx}
    \caption{Overview of classification approaches.}
    \label{tab:overview-classification}
\end{table*}

\begin{table*}
    \centering
    \begin{tabularx}{\textwidth}{l LLLL}
        \toprule
         Team & Base Architecture & Loss Function \& Training Paradigm & Output \& Target Formulation & Inference \& Post-processing \\
         \midrule
         LAiB & ConvNeXtV2-B + FPN & Focal Loss (CenterNet-style) & Heatmaps (2D Gaussian, $128\times128$) & 5-fold ensemble, TTA, soft-argmax peak extraction \\
         LIMED & RadDINO ViT + Heatmap Head & Semi-supervised (Focal Loss \& ASL via CAM pseudo-labels) & Heatmaps (2D Gaussian) & Cascaded pipeline, 5-fold checkpoint merge (greedy soup) \\
         MICLab & CheXagent-2 (3B VLM) & Zero-shot (no fine-tuning) & Bounding boxes (center keypoints) & Semantic filtering, fixed confidence scoring \\
         UQAC & YOLOv8s & Semi-supervised (YOLO losses \& iterative pseudo-labeling) & Bounding boxes (fixed-size from centers) & Cascaded pipeline, high-confidence recovery threshold \\
         \bottomrule
    \end{tabularx}
    \caption{Overview of localization approaches.}
    \label{tab:overview-localization}
\end{table*}

Approaches from the different teams are presented in this section.
Overviews of the different approaches are visible in Tables \ref{tab:overview-classification} and \ref{tab:overview-localization} for an easy comparison of the main strategy points used by the teams.

\subsection{Team CM@MSD}
\begin{figure*}
    \centering
    \includegraphics[width=0.75\linewidth]{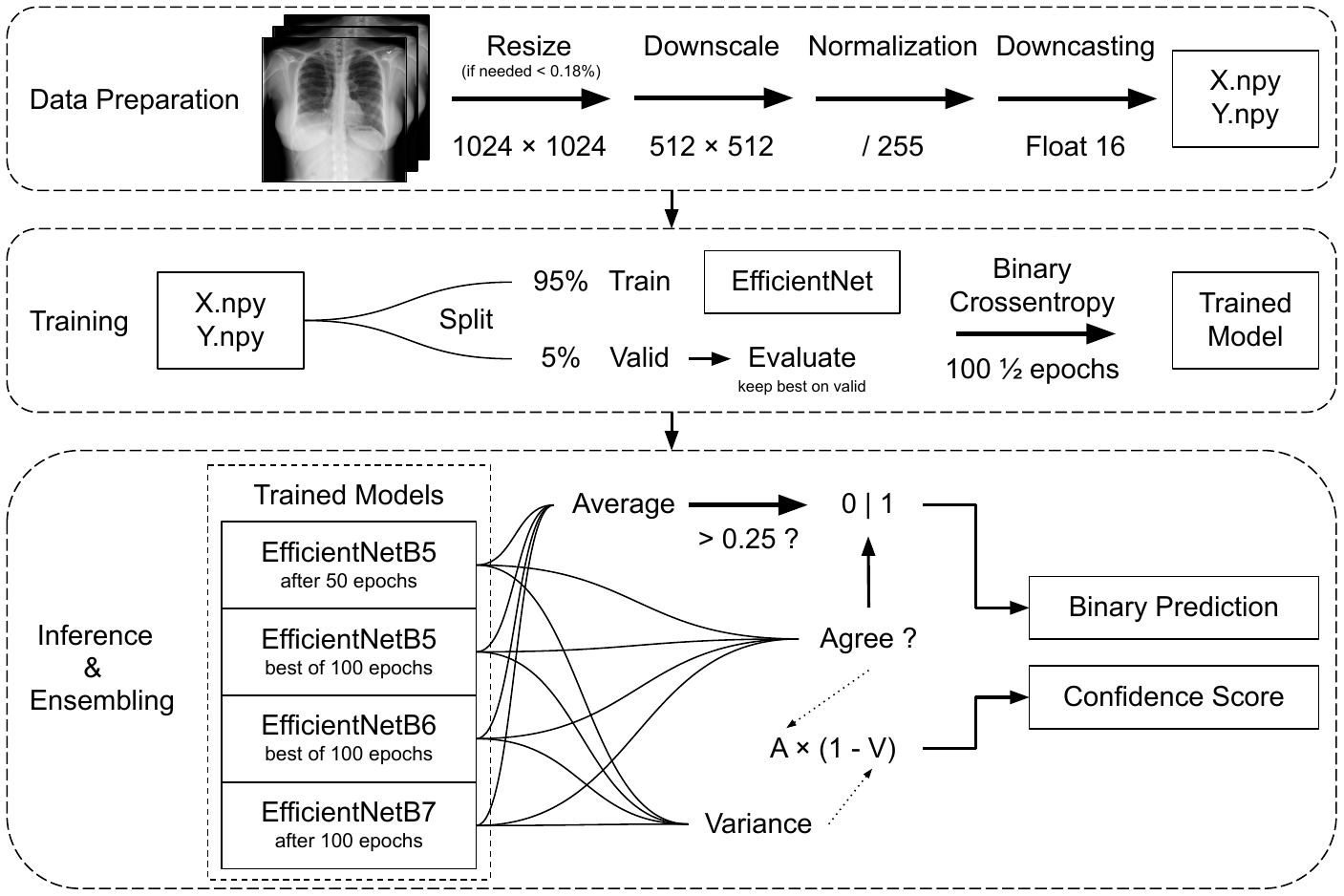}
    \caption{Team CM@MSD's classification pipeline. 
    The approach utilizes an ensemble of EfficientNet architectures (B5, B6, and B7) trained with binary cross-entropy and a batch generator that oversamples positive samples. Inference is performed by averaging probabilistic outputs from four independent models, applying a 0.25 threshold. A final confidence score is generated based on the agreement and variance of the individual model predictions.
    }
    \label{fig:cmmsd}
\end{figure*}
Team CM@MSD (C. Meyer) proposed an approach visible in Figure \ref{fig:cmmsd}.
Source code: \url{https://github.com/Cyril-Meyer/NoduLoCC2026-Classification-Contest}.
\subsubsection{Classification}
Step 1 : Data Preparation.
The original dataset is converted into two NumPy arrays (inputs and labels) to simplify data loading during training. All images are resized to a fixed resolution of $512 \times 512$ pixels using average pooling over $2 \times 2$ blocks. Pixel intensities are normalized to [0,1] range and stored using float16 precision to reduce memory footprint. Images not originally sized at $1024 \times 1024$ are first resized prior to downsampling. Due to an implementation inconsistency in normalization (i.e. the resize function already applies normalization), this introduced a distributional bias correlated with class labels.

Step 2 : Training.
To address class imbalance, this team uses a batch generator that oversamples positive samples, enforcing an approximate 50/50 class ratio within each batch. Models are trained for 100 epochs, corresponding to approximately 50 effective full passes over the dataset using binary crossentropy. Trained architectures are CNN based on the EfficientNet architecture (B5, B6, and B7 variants). For each individual model, 5\% of the training samples are used as a validation set to select the best model. Batch size is adjusted depending on model size and hardware constraints.

Step 3 : Prediction.
At inference time, predictions from four independent models are aggregated. For each input, probabilistic outputs from the models are averaged and then thresholded at 0.25 to obtain a binary prediction. In addition, a confidence score based on the agreement between models and the variance of their predictions is computed. Agreement measures how consistent individual model predictions are with the final binary decision, while variance reflects the dispersion of predicted probabilities. The confidence score is defined as the product of agreement and the inverse of variance.

The ensemble strategy improves generalization while also providing a measure of prediction reliability.

\subsubsection{Training Time and Hardware Used}
Training was done on an NVIDIA RTX 4090 graphics card (24GB VRAM) for a total of 49 hours using the proposed ensemble of EfficientNets. 

\subsection{Team LAiB}
\begin{figure*}
    \centering
    \includegraphics[width=\linewidth]{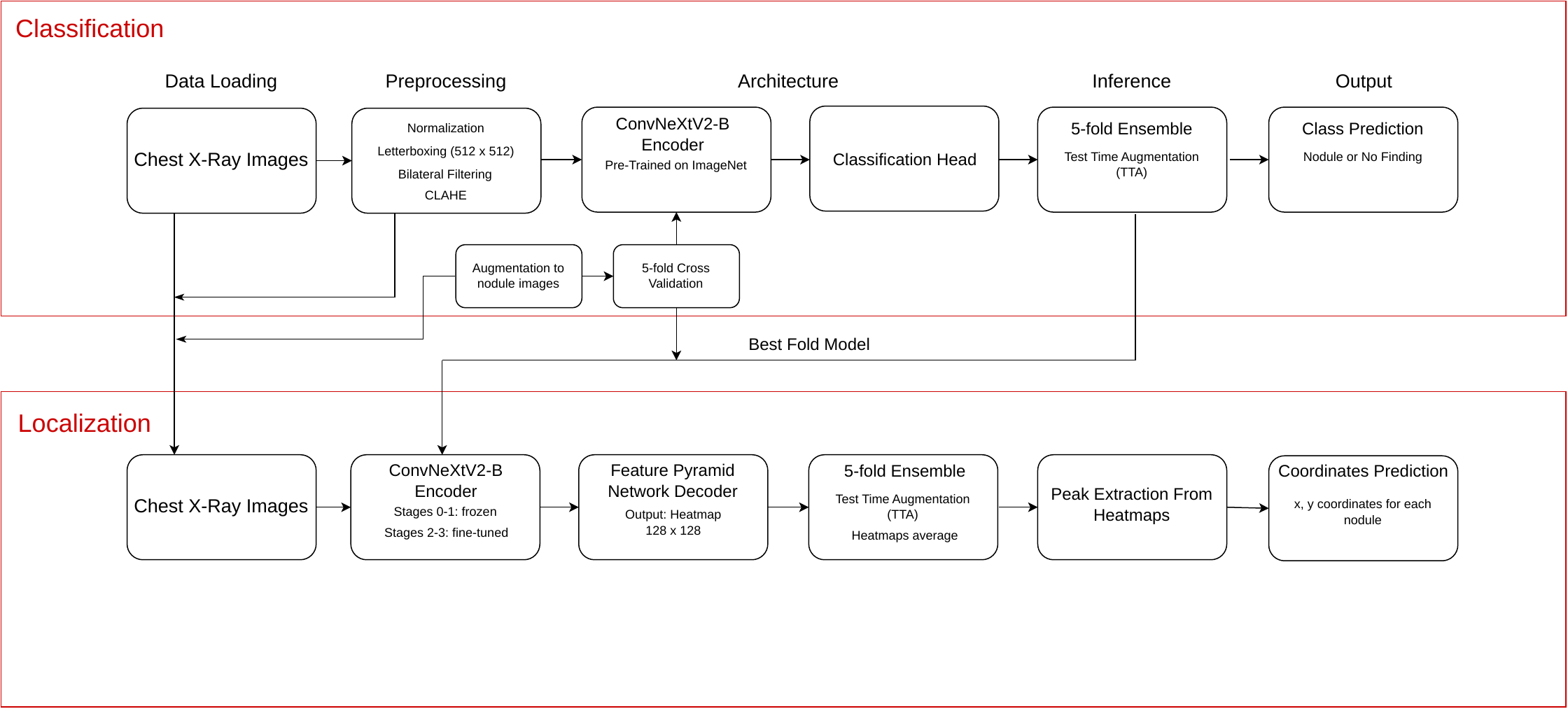}
    \caption{Team LAiB's classification and localization approach.
    The classification phase employs a ConvNeXtV2-B backbone trained under a 5-fold cross-validation strategy, with final predictions generated via a soft ensemble and Test-Time Augmentation (TTA). The localization phase transfers the top-performing encoder to a Feature Pyramid Network (FPN) decoder. This localizer is trained using a CenterNet-style focal loss and outputs heatmaps, which are refined by a soft-argmax operator to extract sub-pixel peak coordinates.
    }
    \label{fig:laib}
\end{figure*}
Team LAiB (C. Bentifece, L. Burattini, I. Marcantoni, A. Sbrollini) proposed approaches visible in Figure \ref{fig:laib}. 
Source code: \url{https://github.com/univpm-br3in/Lung-nodule-challenge_2026}.
\subsubsection{Classification}
Team LAiB's approach treats pulmonary nodule detection as a binary image classification task. 
The architecture employs a ConvNeXtV2-B backbone pre-trained on ImageNet-22k/1k, fine-tuned end-to-end and integrated with a streamlined classification head comprising global average pooling, a dropout layer (rate 0.3), and a single linear unit producing a logit thresholded during inference. 

Team LAiB considered 15,818 chest X-ray images: 113 nodule images from the LIDC dataset and 15,705 no-finding images from the NIH collection, to mitigate class imbalance. 
To address this disparity, they implemented a dual strategy. 
A WeightedRandomSampler enforces balanced 50/50 mini-batches during training, while Mixup regularization ($\alpha=0.2$) synthesizes soft labels from pairs of training samples to mitigate minority-class memorization. 
The loss function is BCEWithLogitsLoss with a positive class weight of 2.0.
Since the batches are already balanced, the elevated weight encodes an explicit clinical prior that penalizes missed nodules twice as heavily as false alarms. 

All images undergo a deterministic pre-processing pipeline cached in RAM via MONAI CacheDataset, computed once before the cross-validation loop. 
Images are loaded at their native bit depth, normalized, and letterboxed to $512\times512$ with zero padding to preserve aspect ratios. 
Noise reduction is performed via bilateral filtering prior to Contrast Limited Adaptive Histogram Equalization.
Critically, the dataset of 15,818 images is cached as a single shared representation reused across all five folds via index-based subsets, avoiding redundant processing and reducing total setup time by a factor of five.
A stochastic augmentation pipeline, including flips, rotations, affine transformations, and photometric perturbations, is then applied at training time exclusively to nodule images, as the no-finding class is already well-represented and does not require artificial oversampling.

The team adopted a 5-fold cross-validation strategy, stratified by class label and grouped by patient ID to prevent any patient-level information from leaking across splits.
Optimization was performed with AdamW using Layer-wise Learning Rate Decay, applying progressively lower learning rates to earlier layers to preserve general visual features while allowing deeper, task-specific layers to specialize freely.
Training efficiency was improved through Automatic Mixed Precision (AMP bfloat16) and torch.compile, yielding approximately 20\% improvement in GPU throughput.

At inference, predictions are generated via a soft ensemble of the five fold models, each contributing four forward passes under test-time augmentations (identity, horizontal flip, and two rotations), for a total of 20 forward passes per image.
The classification threshold of 0.51 was determined through a sweep over Out-Of-Fold (OOF) predictions: after training, each fold model predicts on its own held-out validation set, and concatenating these five sets yields one probability per image always estimated by a model blind to it.
This makes OOF predictions the most rigorous proxy for generalization available without a separate test set, and the threshold maximizing F1 on this distribution is selected for final inference.

\subsubsection{Localization}
This approach treats pulmonary nodule localization as a heatmap regression problem.
To leverage specialized feature representations, the model is not trained from scratch; instead, the team employs a transfer learning strategy using the classification backbone from Task 1.
Specifically, the localization model is initialized with the ConvNeXtV2-B encoder from the top-performing classification fold.
The same pre-processing steps of task 1 are applied, and also in this case augmentation was applied.

To preserve discriminative features while adapting to the localization task, the first two encoder stages remain frozen during training, while the deeper stages are fine-tuned using a reduced learning rate ($5 \times 10-6$).
The decoder utilizes a Feature Pyramid Network (FPN) to aggregate multi-scale feature maps from three distinct encoder stages.
These maps are progressively up-sampled through transposed convolutions, integrated with batch normalization and Gaussian Error Linear Units (GELU) activations.
Skip connections from intermediate encoder layers are fused at each scale via lateral $1 \times 1$ convolutions, enabling the network to combine high-level semantic information with fine-grained spatial details.

The final output is a $128 \times 128$ sigmoid heatmap representing the spatial probability distribution of nodules center.
Target heatmaps are constructed by applying 2D Gaussian distributions at each annotated nodule coordinate.
The spread and amplitude of these Gaussians are modulated by annotation confidence: high-confidence labels produce sharp, narrow peaks, whereas low-confidence labels yield broader distributions.

Optimization is performed using a CenterNet-style focal loss, which concentrates the learning signal on true positive peaks and progressively discounts penalties in the immediate vicinity of the centers, preventing the background gradient from dominating the training process.
The training follows a 5-fold cross-validation strategy.
The team utilizes the AdamW optimizer with differential learning rates for the encoder and decoder, employing a linear warmup followed by a cosine annealing schedule.
Early stopping is monitored via a comprehensive detection metric, mean detection error (MDE) that jointly penalizes localization errors and missed detections.
During each validation epoch, the optimal detection threshold is determined through a grid search to minimize this metric.

At inference time, predictions from the five fold models are combined into an ensemble by averaging their respective heatmaps generated under Test-Time Augmentations (TTA).
Peak coordinates are extracted and refined to sub-pixel precision using a soft-argmax operator within a local spatial window.
Finally, the predicted coordinates are mapped back to the original image space by reversing the letterbox transformations applied during pre-processing.

\subsubsection{Training Time and Hardware Used}
For the classification task, training was completed in $\sim$3.5 hours on an NVIDIA RTX PRO 6000 Blackwell (95 GB VRAM), with an average inference latency of $\sim$93 ms per image.

For the localization task, training was completed in $\sim$ 22 minutes and 47 seconds on an NVIDIA RTX PRO 6000 Blackwell (95 GB VRAM), with an average inference latency of approximately 273 ms per image.

\subsection{Team LIMED}
\begin{figure*}
    \centering
    \includegraphics[width=\linewidth]{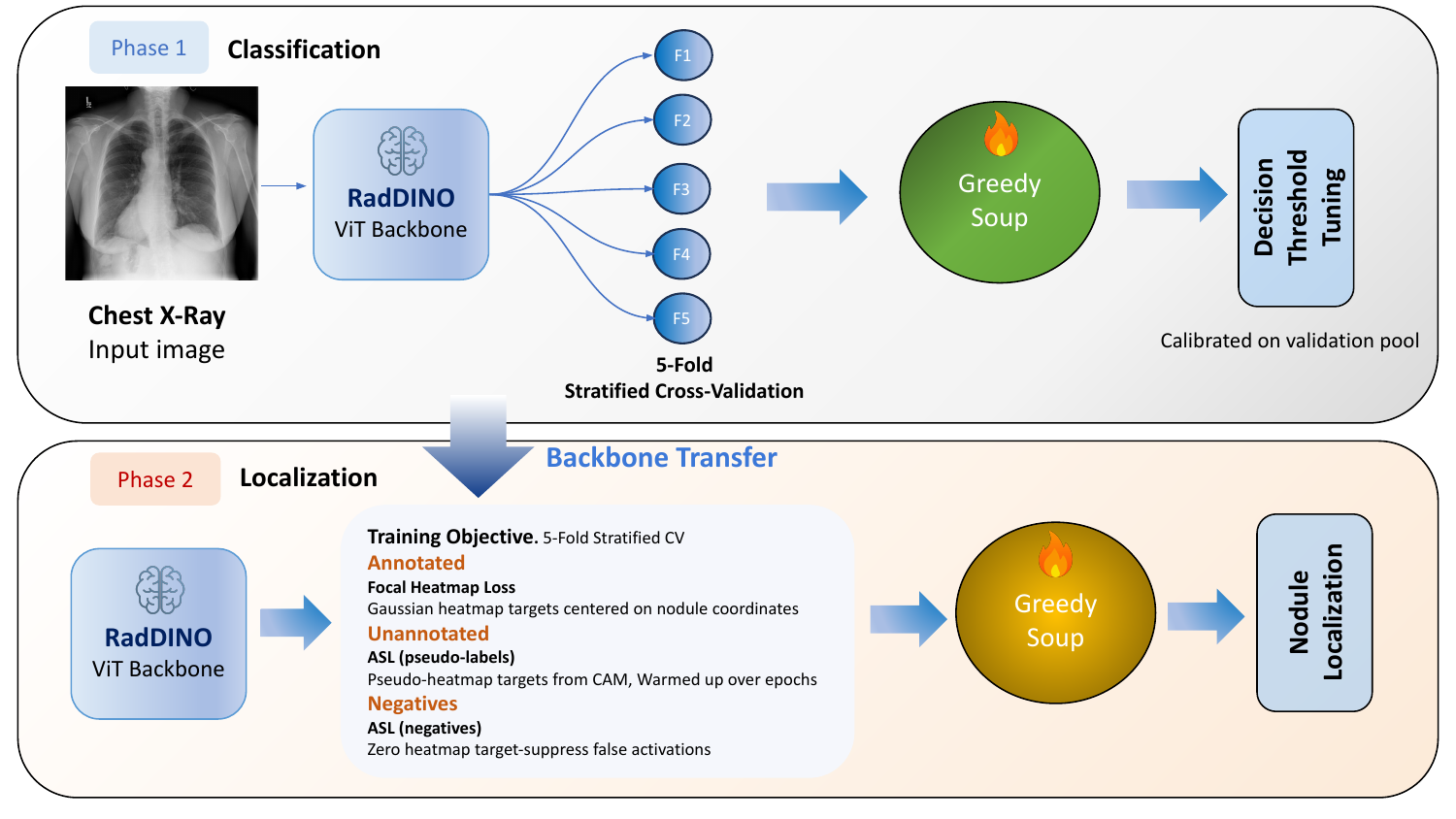}
    \caption{Team LIMED's two-phase cascaded pipeline. Phase 1 (Classification) fine-tunes a RadDINO ViT backbone, merging fold checkpoints via a Greedy Model Soup. Phase 2 (Localization) transfers the frozen backbone to a Heatmap Head trained under multi-objective losses (Focal Heatmap Loss and ASL) and merged via a second Greedy Soup.}
    \label{fig:limed}
\end{figure*}
Team LIMED (F. Idiri, H. Messaoudi, F. Meziane) proposed approaches visible in Figure \ref{fig:limed}. 
Source code: \url{https://github.com/idfahima/NoduLoCC2026---Team-LIMED}.
\subsubsection{Classification}
The first phase focuses on binary classification (Nodule vs. No Finding). This team employs a full RadDINO backbone fine-tuned end-to-end alongside a linear classification head. 
To address the significant class imbalance inherent in the dataset, team LIMED utilizes Asymmetric Sigmoid Loss (ASL) \cite{Ridnik2021} and implement a balanced training sampling strategy (50/50 positive and negative samples). 
The model is trained using a 5-fold stratified cross-validation strategy.
Instead of ensembling at inference time through multiple forward passes, a 5-fold greedy soup strategy is applied \cite{wortsman2022soup}.
The best checkpoints from each of the five folds are merged into a single set of weights that maximizes the held-out F1 score, creating a unified, robust classifier while retaining single-model inference efficiency.

\subsubsection{Localization}
The second phase is dedicated to nodule localization. 
Team LIMED designs a \textit{Heatmap Localizer} initialized with the optimal per-fold checkpoints from the classification phase. 
During this phase, both pretrained RadDINO backbone and the classification head are frozen, allowing only the newly added \textit{HeatmapHead} to learn.
This localizer predicts expected Gaussian heatmaps centered at the annotated nodule coordinates. 
It is optimized using a focal heatmap loss \cite{Lin2020} over annotated samples.
Unannotated positive images are weakly supervised using pseudo-labels generated via Class Activation Mapping (CAM) \cite{Zhou2016}, while negatives and weakly supervised positives use ASL to reduce false positives.
Following a symmetrical 5-fold stratified cross-validation strategy, the phase 2 checkpoints are similarly merged via a greedy fold soup to maximize the localization score on the validation set.

At inference time, the overall pipeline first applies the classifier to an input image.
If the predicted confidence exceeds an optimized presence threshold, the localizer is activated to extract heatmap peaks as the predicted coordinates, yielding precise localizations with controlled false-positive rates.

\subsubsection{Training Time and Hardware Used}
The models were developed and trained using an NVIDIA RTX 4090 GPU with 24 GB of VRAM. 
For the Phase 1 classification model, training on 5 folds for 30 epochs per fold took approximately 10 hours. 
For the Phase 2 Localization model, training the folds for 40 epochs per fold took roughly 5 hours. 
The entire model convergence pipeline, including inference threshold calibrations and greedy fold soup generation, was completed in under 15 hours on a single GPU workstation.

\subsection{Team MICLab}
\begin{figure*}
    \centering
    \includegraphics[width=0.90\linewidth]{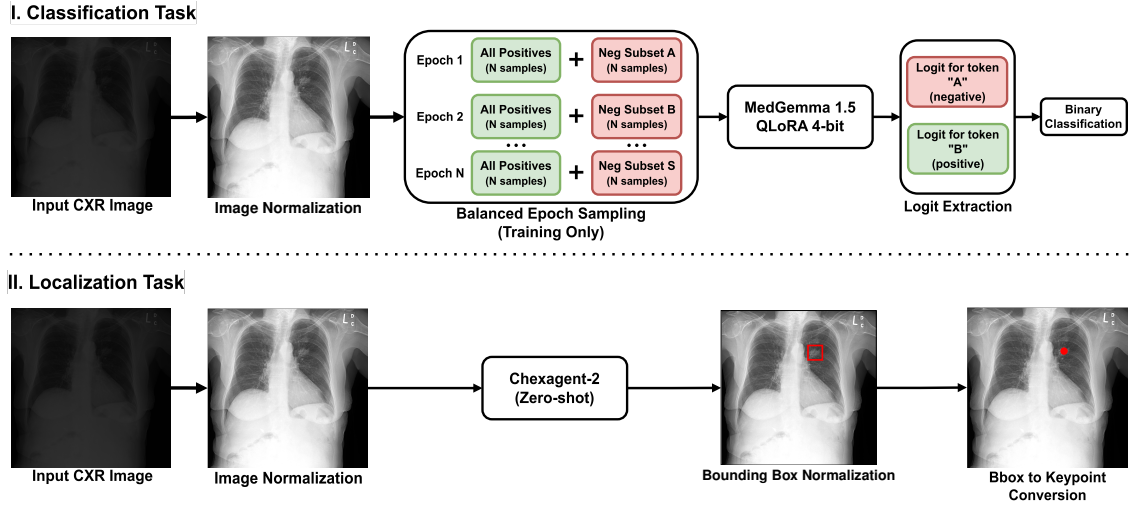}
    \caption{Team MICLab's Vision-Language Model (VLM) pipelines.
    The classification pipeline fine-tunes MedGemma 1.5 using QLORA 4-bit precision and a rotating balanced epoch sampler, formulating the detection task as constrained next-token prediction. The localization phase relies on a zero-shot approach using CheXagent-2. Spatial predictions are derived by converting the model's normalized bounding box outputs into single center keypoints.}
    \label{fig:miclab}
\end{figure*}

Team MICLab (K.A.O. Aguilar, D. Carmo, L. Rittner) proposed approaches visible in Figure \ref{fig:miclab}. 
Source code: \url{https://github.com/MICLab-Unicamp/nodulocc_challenge}.
\subsubsection{Classification}

For the classification task, the goal of this team was to evaluate the transferability of modern medical Vision-Language Models (VLMs) under the challenge's constraints of severe class imbalance. 
After evaluating several zero-shot and fine-tuned models, MedGemma 1.5 \cite{sellergren2025medgemma} demonstrated the best precision-recall tradeoff. 
MedGemma 1.5 is a 4B-parameter VLM utilizing a SigLIP image encoder \cite{siglip-2023} and Gemma 3 \cite{gemma_2025} as a decoder-only transformer.

To adapt the model while preventing overfitting and reducing VRAM requirements, team MICLab employed QLoRA (Quantized Low-Rank Adaptation) fine-tuning, quantizing the base weights to \mbox{4-bit} NF4 precision. 
This preserved the model's pre-trained medical knowledge while updating only a small set of adapter weights.

Team MICLab formulated the classification problem as a constrained next-token prediction task. 
The model was prompted to produce a single-character answer corresponding to the class label: ''A`` for healthy and ''B`` for nodule. 
During training and evaluation, they extracted the logits associated with these two tokens at the first supervised answer position, yielding a two-dimensional output vector for each sample.
These two logits naturally define a binary classification decision, mapped to classes 0 and 1, which makes it possible to optimize the model directly with weighted cross-entropy loss and to compute standard classification metrics such as accuracy, precision, recall, specificity, F1 score, balanced accuracy, MCC, ROC-AUC, PR-AUC, and Brier score without requiring free-text generation parsing.

To directly tackle the extreme 95:5 (healthy:nodule) class imbalance, this team implemented two complementary strategies:
\begin{enumerate}
    \item Weighted Cross-Entropy Loss: Positive samples were assigned a weight of 20, heavily penalizing missed nodules relative to false alarms.
    \item Rotating Balanced Epoch Sampler: Each training epoch included all positive samples alongside a rotating, non-overlapping subset of negative samples of equal size.
    This maintained a 1:1 class ratio per epoch while ensuring the model eventually evaluated all negative samples without data waste.
\end{enumerate}

The model was trained using an effective batch size of 16 (via gradient accumulation), the AdamW optimizer, and a linear warmup-decay learning rate schedule.
This design allowed to retain the generative VLM formulation while making the optimization and evaluation pipeline equivalent to a standard binary classifier operating on two logits.

Image normalization, detailed in the Localization task section, was applied identically to all inputs before inference.
\subsubsection{Localization}

The primary challenge in the localization task was the extremely limited number of annotated keypoints (179 keypoints across 113 images) coupled with a significant distribution shift between training and test sets. Initial attempts to fine-tune various VLMs and CNNs resulted in rapid overfitting.

To circumvent this, the team opted for a zero-shot approach using CheXagent-2 \cite{chexagent-2024}, a 3B-parameter VLM specialized for chest X-ray interpretation. 
By avoiding fine-tuning, they leveraged its robust pre-trained knowledge base derived from large-scale medical datasets without explicitly biasing the model to the small training subset.

A key component of both classification and localization pipelines is image normalization before inference. 
This normalization was essential because, without DICOM windowing metadata, direct rescaling of the raw int16 images from the LIDC-IDRI dataset would yield low-contrast inputs and degrade VLM inference.
Each input image was converted to a single grayscale channel, clipped to fixed lower and upper pixel-value percentiles, and then rescaled to the \([0,255]\) range before being converted back to RGB for model compatibility.
This normalization step was applied uniformly to different image bit depths and helped suppress outliers such as burnt-in annotations and detector artifacts while improving contrast consistency across studies.

CheXagent-2 produces localization outputs in bounding-box format using normalized coordinates.
The team parsed these predicted bounding boxes, filtered them according to the semantic label associated with each region (retaining only labels consistent with lung nodules when desired), and then converted the normalized \([0,100]\) coordinates back to image-space pixel coordinates using the original image width and height.
Since the challenge submission format requires point annotations rather than bounding boxes, they transformed each predicted box into a single point by taking the center of the box.
These center points were then exported as the final \((x,y)\) localization predictions, with a confidence score of 1.0 for each retained detection.

\subsubsection{Training Time and Hardware Used}
- Hardware (used for both tasks): Single NVIDIA GeForce RTX 4090 GPU (24GB VRAM).
- Classification Task Training Time: Approximately 35 hours (using QLoRA fine-tuning).
- Localization Task Training Time: N/A (0 hours). Due to the challenges of overfitting on a small annotated dataset, this model was used strictly in a zero-shot inference capacity.

\subsection{Team UQAC}
\begin{figure*}
    \centering
    \includegraphics[width=0.8\linewidth]{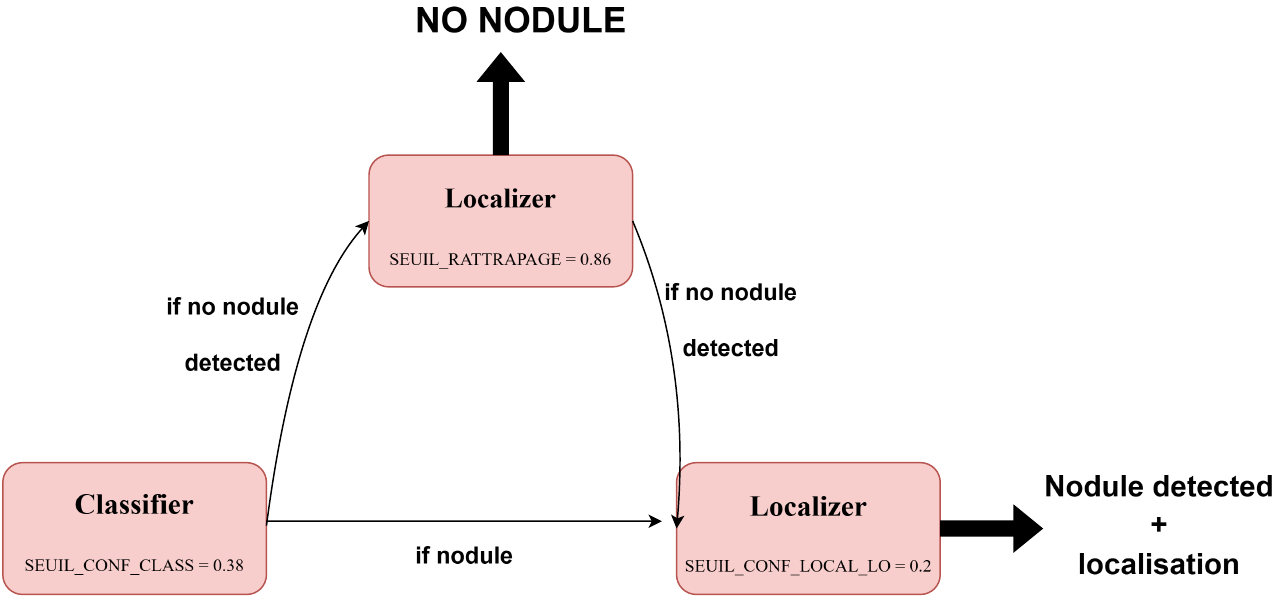}
    \caption{Team UQAC's cascaded dual-reader decision pipeline.
Phase 1 (Classification) utilizes an EfficientNetV2-S architecture enhanced with a Multi-scale Embedding Attention Mechanism, trained using focal loss and weighted random sampling. Phase 2 (Localization) employs a YOLOv8s detector trained via iterative semi-supervised pseudo-labeling, with final predictions determined by a cascaded process relying on jointly optimized classification and high-confidence recovery thresholds.}
    \label{fig:uqac}
\end{figure*}
Team UQAC (R. Amigon, C. Bardin, M. Boehm, K. Bouchard, M. Lahargoue, L. Morales, L. Zipper) proposed approaches visible in Figure \ref{fig:uqac}. 
Team UQAC source code: \url{https://github.com/Clement1180/Nodule_detection}.
\subsubsection{Classification}
Team UQAC models detection as a binary classification problem $y \in \{0,1\}$.
Each radiograph $x \in \mathbb{R}^{512\times512}$ is transformed into a three-channel tensor $\tilde{x} = [x_r,x_s,x_c]$, where $x_r$ is min–max normalized intensity, $x_s$ is a Sparse Edge-Preserving Enhancement (SEPE) \cite{Dhayalini2025}: 
\[x_s = \mathcal{M}_{\text{Sobel}}(x) \odot (x + \lambda \Delta x),\]
with $\mathcal{M}_{\text{Sobel}}$ an adaptive gradient mask and $\Delta$ the Laplacian operator, and $x_c$ is obtained via CLAHE \cite{SIDHU2023104477}. 
The model utilizes EfficientNetV2-S, pre-trained on ImageNet-1K, to extract feature maps at four different resolution levels \cite{tan2021efficientnetv2smallermodelsfaster}. 
These are fused by a Multi-scale Embedding Attention Mechanism \cite{Dhayalini2025}: 
\[Attn(Q,K,V) = Softmax \left( \frac{QK^\mathsf{T}}{\sqrt{d}}\right)V,\]
where $(Q,K,V)$ are learned projections of spatially aligned feature maps. 
The fused representation is then processed by an Adaptive Average Pooling layer followed by a fully connected layer of 128 units with ReLU activation, a Dropout layer $(p = 0.5)$, and a final linear layer producing the class probability.

Training minimizes the focal loss combined with Weighted Random Sampling to expose the minority class more frequently. 
Optimization uses AdamW with cosine annealing. 
The model achieves 88.62\% recall, increasing to 93.6\% after threshold tuning. 
\subsubsection{Localization}
Localization is formulated as object detection using YOLOv8s, initialized from VinBigData pretraining \cite{nguyen2022vindrcxropendatasetchest}.
Pre-processing applies percentile clipping, thoracic masking, CLAHE, and Gaussian smoothing.

Input images undergo a standardized preprocessing pipeline to reduce variability and enhance anatomical structures.
Radiographs are converted to grayscale and normalized using 1st–99th percentile clipping. 
A smoothed elliptical thoracic mask is applied.
Then, local contrast is enhanced using CLAHE (clip limit 2.0, $8 \times 8$ grid), followed by a $3 \times 3$ Gaussian filter for denoising. 
Images are converted to RGB and resized $x \in \mathbb{R}^{512\times512}$.

To address limited annotations, team UQAC employed an iterative semi-supervised pseudo-labeling strategy.
The model is first trained on labeled images, where point annotations $(x,y)$ are converted into fixed-size bounding boxes.
At each iteration $t$, pseudo-labels are generated on unlabeled positive images and filtered using confidence thresholds and Non-Maximum Suppression(NMS).

\[\hat{\mathcal{Y}}_u^{(t)} = \text{NMS} (\{b \mid p(b) > \tau_p\})\]
and incorporated only if $\mathrm{mAP}_{50} > 0.35$, preventing error propagation. 
The model is then retrained on the augmented dataset, progressively improving recall. 
Training minimizes standard YOLO losses:
\[
\mathcal{L}=\mathcal{L_{\text{box}}}+\mathcal{L_{\text{obj}}}+\mathcal{L_{\text{cls}}}\text{.}
\]

On the annotated subset, the model achieves a precision of 0.625, a recall of 0.769, and an F1-score of 0.690, with an average center-to-center localization error of 6.79 pixels.

Pipeline. To maximize sensitivity while controlling false positives, this team defines a cascaded dual-reader decision process combining the classifier and the detector. 
Let $p_c(x) \in [0, 1]$ denote the classifier probability for image $x$, and let $p_d (b|x)$ denote the confidence score of a detected bounding box $b$ produced by YOLOv8s.

The final image-level prediction $\hat{y}$ is defined as:
\[
\hat{y} = 
\begin{cases} 
1 & \text{if } p_c(x) > \tau_c, \\
1 & \text{if } p_c(x) \leq \tau_c \land \max\limits_{b \in \mathcal{B}(x)} p_d(b|x) > \tau_r, \\
0 & \text{otherwise},
\end{cases}
\]
where $\mathcal{B}(x)$ denotes the set of predicted bounding boxes for image $x,\tau_c=0.38$ is the classification threshold, and $\tau_r=0.86$ is a high-confidence recovery threshold.

For images such that $\hat{y}= 1$, final localization is performed by selecting bounding boxes with $p_d(b|x) > \tau_l$, where $\tau_l = 0.2$, ensuring high recall at the spatial level.

The thresholds $(\tau_c, \tau_r)$ are jointly optimized via grid search on a balanced validation set $(n = 1000)$ to maximize the global F1-score to 0.6897.

\subsubsection{Training Time and Hardware Used}
For the classification task, training time is 1.5h on an NVIDIA RTX 4070 Super.
For the localization task, Training time is 40 minutes on an NVIDIA RTX 4070 Super.

\section{Results}

\FloatBarrier
\subsection{Classification}

\begin{figure}
    \centering
    \includegraphics[width=\linewidth]{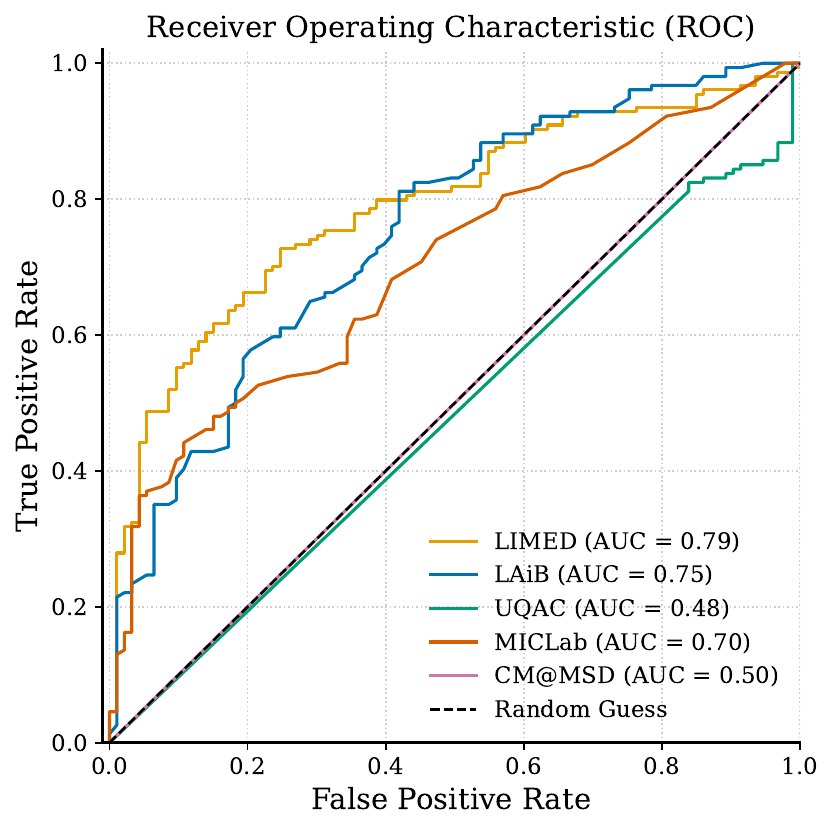}
    \caption{ROC Curves for the classification task on the JSRT test set.}
    \label{fig:roc_curves}
\end{figure}

\begin{table*}[ht]
    \centering
    \caption{Classification performance on the JSRT test set. Square brackets show the 95th percentile confidence using bootstrapping with 1000 iterations.}
    \begin{tabular}{l cccc c}
        \toprule
        Team    & Balanced Accuracy     & AUC-ROC                   & Recall    & Specificity   & Precision \\
        \midrule
        LIMED   & \B{0.72} [0.67, 0.78]  & \B{0.79} [0.73, 0.85]    & 0.66      & 0.78          & 0.84      \\
        LAiB    & 0.67 [0.61, 0.72]      & 0.75 [0.69, 0.82]        & 0.86      & 0.47          & 0.73      \\
        MICLab  & 0.59 [0.56, 0.63]      & 0.70 [0.64, 0.76]        & 0.21      & \B{0.97}      & \B{0.92}  \\
        UQAC    & 0.54 [0.50, 0.58]      & 0.48 [0.43, 0.52]        & 0.96      & 0.12          & 0.64      \\
        CM@MSD  & 0.50 [0.50, 0.50]      & 0.50 [0.50, 0.50]        & \B{1.00}  & 0.00          & 0.62      \\
        \bottomrule
    \end{tabular}
    \label{tab:classif_jsrt}
\end{table*}

Table \ref{tab:classif_jsrt} shows the results for the classification task for the JSRT test set.
We can see that the proposed methods struggle to balance recall and specificity with teams CM@MSD and UQAC having a very high recall with low specificity whereas team MICLab's approach shows the inverse trend.
Notably, team CM@MSD proposed a method which fails to generalize to our provided test data. 
We have confirmed an issue in the normalization step during their pre-treatment, results shown in this table do not reflect the quality of the proposed method.
Due to challenge conditions, we could not allow for code changes past the deadline neither for inference nor for training.
Team LAiB and team LIMED's approaches stand out as better performing methods for this task. 
Among these two approaches, LAiB shows a better recall than LIMED, which itself leads in other metrics.  
LIMED shows the best overall performance showing at least a score of 0.6 in each metric.

While team LIMED achieved the highest overall AUC-ROC of 0.79 compared to Team LAiB’s 0.75, we performed statistical testing to determine if this performance margin was significant. 
Because both models were evaluated on the same test set (JSRT), we utilized DeLong’s test for two correlated ROC curves \cite{DeLong1988-zz}. 
The analysis yielded a p-value of $p=0.32$. 
Assuming a standard significance level of $\alpha=0.05$, the difference in discriminative performance between LIMED and LAiB is not statistically significant, indicating that the dataset size limits our ability to declare a definitive absolute winner between the top two approaches.

Figure \ref{fig:roc_curves} shows the full ROC curves for the test set across all teams. 
We observe LIMED leads in ROC from a false positive rate of 0 to about 0.4, where LAiB starts to lead with a slight edge that starts to grow at a false positive rate from 0.7 until 1. 
While both methods are very similar at false positive rates between 0.4 and 0.7, team LIMED's proposition leads in low false positive rate scenarios making it overall a more exploitable solution, this is also indicated by the AUC values of 0.79 for LIMED and 0.75 for LAiB.

\subsection{Localization}

\begin{table*}
    \centering
    \caption{Localization results for the JSRT test set per team. Coverage is defined as images where the correct number of nodules was found in the image. Distances are measured in mm on covered images.}
    \begin{tabular}{l cccc cc}
        \toprule
        & & \multicolumn{4}{c}{Distance (mm)} & \\
        \cmidrule(lr){3-6}
        Team    & Coverage ($\uparrow$) & Mean      & Median    & Min       & Max           & Standard Deviation    \\
        \midrule
        MICLab  & \B{0.53}  & \B{19.02} & \B{12.83} & 1.44      & \B{171.04}    & \B{25.46}             \\
        UQAC    & 0.17      & 86.16     & 80.44     & \B{0.69}  & 273.97        & 76.60                 \\
        LIMED   & 0.05      & 99.88     & 106.79    & 4.47      & 207.60        & 62.00                 \\
        LAiB    & 0.00      & -         & -         & -         & -             & -                     \\
        \bottomrule
    \end{tabular}
    \label{tab:loc_jsrt}
\end{table*}

For the localization task we define coverage as representing the proportion of images where the number of predicted nodules is the same as the number of nodules present in the image. 
The distance metrics are measured in mm on covered images. 
Table \ref{tab:loc_jsrt} shows these metrics for our test set. 
Notably, team LAiB's solution provided zero images for which it predicted the correct number of nodules. 
Their solution overestimates the number of nodules present in the radiographs.
Propositions from teams UQAC and LIMED appear similar, with UQAC leading in coverage, mean, median and minimum distances.
Both of these methods lack in terms of coverage.
We observe that Team MICLab's solution shows the highest coverage along with lowest mean, median and maximum distance and standard deviation between distances, showing a clear gap in performance between this method and others.
It needs to be reminded that a nodule is defined as a growth of less than 30mm in diameter, meaning that any distance over 15mm points outside of the actual nodule. 
As none of the proposed methods show a mean or median distance of less than 15mm, we show that the proposed task is difficult to solve perfectly.

\section{Discussion}

Team LIMED and LAiB outperformed for the classification task due to shared metric optimization strategies. 
Namely, both teams employed 5-fold stratified cross-validation, a merging technique for inference, and decision threshold tuning.
These methods were not present in the other classification proposals, showing their positive impact on the resulting models.

Interestingly, comparing the ROC curves for LAiB and LIMED's solution, we noted a p-value of 0.32. This indicates that, at least for our test set, the proposed solutions are not statistically distinct. This in turn shows that the similarities in concepts between the methods outweigh their difference in implementation choices.

Team MICLab's proposal of using a VLM for the localization task outperformed the pseudo-labeling employed by team UQAC and the heatmap-based methods proposed by teams LIMED and LAiB, showing that using a zero-shot VLM can lead to interesting results in an extremely data-constrained task.
It needs to be noted that although MICLab's proposed solution is a VLM trained on 32 source datasets composed of 8.5M images, the model was not trained on the JSRT dataset according to CheXagent's authors \cite{chexagent}. 

\section{Conclusion}
We proposed NoduLoCC2026, a challenge focused on detecting and localizing nodules in chest X-ray images sourced from the NIH ChestX-ray14 and the LIDC-IDRI datasets.
In this challenge we have received submissions from five teams across Europe, Africa, and the Americas, showing the impact and interest of the biomedical image analysis community for this task.
The challenge paradigm allows for creative collaboration as a competition, leading to differing proposals for taking on the same problem.
The different propositions were tested on the JSRT dataset, showing generalization capabilities for images from unrelated patients, sensors, hospitals, and dates.

All participants methods are openly available on their respective github repositories.

\FloatBarrier
\balance
\bibliographystyle{plain}
\bibliography{bibliography}

@article{raddino2025,
  title = {Exploring scalable medical image encoders beyond text supervision},
  volume = {7},
  ISSN = {2522-5839},
  url = {http://dx.doi.org/10.1038/s42256-024-00965-w},
  DOI = {10.1038/s42256-024-00965-w},
  number = {1},
  journal = {Nature Machine Intelligence},
  publisher = {Springer Science and Business Media LLC},
  author = {Pérez-García,  Fernando and Sharma,  Harshita and Bond-Taylor,  Sam and Bouzid,  Kenza and Salvatelli,  Valentina and Ilse,  Maximilian and Bannur,  Shruthi and Castro,  Daniel C. and Schwaighofer,  Anton and Lungren,  Matthew P. and Wetscherek,  Maria Teodora and Codella,  Noel and Hyland,  Stephanie L. and Alvarez-Valle,  Javier and Oktay,  Ozan},
  year = {2025},
  month = jan,
  pages = {119–130}
}

@inproceedings{Ridnik2021,
  title = {Asymmetric Loss For Multi-Label Classification},
  url = {http://dx.doi.org/10.1109/ICCV48922.2021.00015},
  DOI = {10.1109/iccv48922.2021.00015},
  booktitle = {2021 IEEE/CVF International Conference on Computer Vision (ICCV)},
  publisher = {IEEE},
  author = {Ridnik,  Tal and Ben-Baruch,  Emanuel and Zamir,  Nadav and Noy,  Asaf and Friedman,  Itamar and Protter,  Matan and Zelnik-Manor,  Lihi},
  year = {2021},
  month = oct,
  pages = {82–91}
}

@inproceedings{Zhou2016,
  title = {Learning Deep Features for Discriminative Localization},
  url = {http://dx.doi.org/10.1109/CVPR.2016.319},
  DOI = {10.1109/cvpr.2016.319},
  booktitle = {2016 IEEE Conference on Computer Vision and Pattern Recognition (CVPR)},
  publisher = {IEEE},
  author = {Zhou,  Bolei and Khosla,  Aditya and Lapedriza,  Agata and Oliva,  Aude and Torralba,  Antonio},
  year = {2016},
  month = jun,
  pages = {2921–2929}
}

@article{Lin2020,
  title = {Focal Loss for Dense Object Detection},
  volume = {42},
  ISSN = {1939-3539},
  url = {http://dx.doi.org/10.1109/TPAMI.2018.2858826},
  DOI = {10.1109/tpami.2018.2858826},
  number = {2},
  journal = {IEEE Transactions on Pattern Analysis and Machine Intelligence},
  publisher = {Institute of Electrical and Electronics Engineers (IEEE)},
  author = {Lin,  Tsung-Yi and Goyal,  Priya and Girshick,  Ross and He,  Kaiming and Dollar,  Piotr},
  year = {2020},
  month = feb,
  pages = {318–327}
}

@InProceedings{wortsman2022soup,
  title = 	 {Model soups: averaging weights of multiple fine-tuned models improves accuracy without increasing inference time},
  author =       {Wortsman, Mitchell and Ilharco, Gabriel and Gadre, Samir Ya and Roelofs, Rebecca and Gontijo-Lopes, Raphael and Morcos, Ari S and Namkoong, Hongseok and Farhadi, Ali and Carmon, Yair and Kornblith, Simon and Schmidt, Ludwig},
  booktitle = 	 {Proceedings of the 39th International Conference on Machine Learning},
  pages = 	 {23965--23998},
  year = 	 {2022},
  editor = 	 {Chaudhuri, Kamalika and Jegelka, Stefanie and Song, Le and Szepesvari, Csaba and Niu, Gang and Sabato, Sivan},
  volume = 	 {162},
  series = 	 {Proceedings of Machine Learning Research},
  month = 	 {17--23 Jul},
  publisher =    {PMLR},
  url = 	 {https://proceedings.mlr.press/v162/wortsman22a.html}
}

@article{sellergren2025medgemma,
  title={MedGemma Technical Report},
  author={Sellergren, Andrew and Kazemzadeh, Sahar and Jaroensri, Tiam and Kiraly, Atilla and Traverse, Madeleine and Kohlberger, Timo and Xu, Shawn and Jamil, Fayaz and Hughes, Cían and Lau, Charles and others},
  journal={arXiv preprint arXiv:2507.05201},
  year={2025}
}

@article{chexagent-2024,
  title={CheXagent: Towards a Foundation Model for Chest X-Ray Interpretation},
  author={Chen, Zhihong and Varma, Maya and Delbrouck, Jean-Benoit and Paschali, Magdalini and Blankemeier, Louis and Veen, Dave Van and Valanarasu, Jeya Maria Jose and Youssef, Alaa and Cohen, Joseph Paul and Reis, Eduardo Pontes and Tsai, Emily B. and Johnston, Andrew and Olsen, Cameron and Abraham, Tanishq Mathew and Gatidis, Sergios and Chaudhari, Akshay S and Langlotz, Curtis},
  journal={arXiv preprint arXiv:2401.12208},
  url={https://arxiv.org/abs/2401.12208},
  year={2024}
}

@misc{siglip-2023,
  title={Sigmoid Loss for Language Image Pre-Training},
  author={Xiaohua Zhai and Basil Mustafa and Alexander Kolesnikov and Lucas Beyer},
  year={2023},
  eprint={2303.15343},
  archivePrefix={arXiv},
  primaryClass={cs.CV},
  url={https://arxiv.org/abs/2303.15343},
}

@misc{gemma_2025,
  title={Gemma 3},
  url={https://goo.gle/Gemma3Report},
  publisher={Kaggle},
  author={Gemma Team},
  year={2025}
}

@Article{Dhayalini2025,
author={M, Dhayalini
and B, Revathi alias Ponmozhi},
title={Multi-phase deep learning framework with Multiscale Adaptive Swin Transformer and embedding attention for precision lung nodule detection and classification},
journal={Scientific Reports},
year={2025},
month={Dec},
day={13},
volume={16},
number={1},
pages={1652},
issn={2045-2322},
doi={10.1038/s41598-025-31147-2},
url={https://doi.org/10.1038/s41598-025-31147-2}
}

@misc{nguyen2022vindrcxropendatasetchest,
      title={VinDr-CXR: An open dataset of chest X-rays with radiologist's annotations}, 
      author={Ha Q. Nguyen and Khanh Lam and Linh T. Le and Hieu H. Pham and Dat Q. Tran and Dung B. Nguyen and Dung D. Le and Chi M. Pham and Hang T. T. Tong and Diep H. Dinh and Cuong D. Do and Luu T. Doan and Cuong N. Nguyen and Binh T. Nguyen and Que V. Nguyen and Au D. Hoang and Hien N. Phan and Anh T. Nguyen and Phuong H. Ho and Dat T. Ngo and Nghia T. Nguyen and Nhan T. Nguyen and Minh Dao and Van Vu},
      year={2022},
      eprint={2012.15029},
      archivePrefix={arXiv},
      primaryClass={eess.IV},
      url={https://arxiv.org/abs/2012.15029}, 
}

@article{SIDHU2023104477,
title = {Segmentation of retinal blood vessels by a novel hybrid technique- Principal Component Analysis (PCA) and Contrast Limited Adaptive Histogram Equalization (CLAHE)},
journal = {Microvascular Research},
volume = {148},
pages = {104477},
year = {2023},
issn = {0026-2862},
doi = {https://doi.org/10.1016/j.mvr.2023.104477},
url = {https://www.sciencedirect.com/science/article/pii/S0026286223000031},
author = {R.K. Sidhu and Jainy Sachdeva and D. Katoch},
}

@misc{tan2021efficientnetv2smallermodelsfaster,
      title={EfficientNetV2: Smaller Models and Faster Training}, 
      author={Mingxing Tan and Quoc V. Le},
      year={2021},
      eprint={2104.00298},
      archivePrefix={arXiv},
      primaryClass={cs.CV},
      url={https://arxiv.org/abs/2104.00298}, 
}

@article{jsrt,
author = {{Shiraishi, Junji and Katsuragawa, Shigehiko and Ikezoe, Junpei and Matsumoto, Tsuneo and Kobayashi, Takeshi and Komatsu, Ken-ichi and Matsui, Mitate and Fujita, Hiroshi and Kodera, Yoshie and Doi, Kunio}},
title = {{Development of a Digital Image Database for Chest Radiographs With and Without a Lung Nodule}},
journal = {American Journal of Roentgenology},
volume = {174},
number = {1},
pages = {71-74},
year = {2000},
doi = {10.2214/ajr.174.1.1740071},
note ={PMID: 10628457},
URL = {https://doi.org/10.2214/ajr.174.1.1740071},
eprint = {https://doi.org/10.2214/ajr.174.1.1740071}
}

@inproceedings{cxr8,
  title={Chestx-ray8: Hospital-scale chest x-ray database and benchmarks on weakly-supervised classification and localization of common thorax diseases},
  author={Wang, Xiaosong and Peng, Yifan and Lu, Le and Lu, Zhiyong and Bagheri, Mohammadhadi and Summers, Ronald M},
  booktitle={Proceedings of the IEEE conference on computer vision and pattern recognition},
  pages={2097--2106},
  year={2017},
  howpublished = "\url{https://openaccess.thecvf.com/content_cvpr_2017/papers/Wang_ChestX-ray8_Hospital-Scale_Chest_CVPR_2017_paper.pdf}"
}

@MISC{lidc-idri,
  title     = "{Data From LIDC-IDRI}",
  author    = "Armato, III, Samuel G and McLennan, Geoffrey and Bidaut, Luc and
               McNitt-Gray, Michael F and Meyer, Charles R and Reeves, Anthony
               P and Zhao, Binsheng and Aberle, Denise R and Henschke, Claudia
               I and Hoffman, Eric A and Kazerooni, Ella A and MacMahon, Heber
               and Van Beek, Edwin J R and Yankelevitz, David and Biancardi,
               Alberto M and Bland, Peyton H and Brown, Matthew S and
               Engelmann, Roger M and Laderach, Gary E and Max, Daniel and
               Pais, Richard C and Qing, David P Y and Roberts, Rachael Y and
               Smith, Amanda R and Starkey, Adam and Batra, Poonam and
               Caligiuri, Philip and Farooqi, Ali and Gladish, Gregory W and
               Jude, C Matilda and Munden, Reginald F and Petkovska, Iva and
               Quint, Leslie E and Schwartz, Lawrence H and Sundaram, Baskaran
               and Dodd, Lori E and Fenimore, Charles and Gur, David and
               Petrick, Nicholas and Freymann, John and Kirby, Justin and
               Hughes, Brian and Casteele, Alessi Vande and Gupte, Sangeeta and
               Sallam, Maha and Heath, Michael D and Kuhn, Michael H and
               Dharaiya, Ekta and Burns, Richard and Fryd, David S and
               Salganicoff, Marcos and Anand, Vikram and Shreter, Uri and
               Vastagh, Stephen and Croft, Barbara Y and Clarke, Laurence P",
  publisher = "The Cancer Imaging Archive",
  year      =  2015
}

@article{macmahon2017guidelines,
  title={Guidelines for management of incidental pulmonary nodules detected on CT images: from the Fleischner Society 2017},
  author={MacMahon, Heber and Naidich, David P and Goo, Jin Mo and Lee, Kyung Soo and Leung, Ann NC and Mayo, John R and Mehta, Atul C and Ohno, Yoshiharu and Powell, Charles A and Prokop, Mathias and others},
  journal={Radiology},
  volume={284},
  number={1},
  pages={228--243},
  year={2017},
  publisher={Radiological Society of North America}
}

@article{national2011reduced,
  title={Reduced lung-cancer mortality with low-dose computed tomographic screening},
  author={National Lung Screening Trial Research Team},
  journal={New England Journal of Medicine},
  volume={365},
  number={5},
  pages={395--409},
  year={2011},
  publisher={Mass Medical Soc}
}

@article{brenner2007computed,
  title={Computed tomography—an increasing source of radiation exposure},
  author={Brenner, David J and Hall, Eric J},
  journal={New England journal of medicine},
  volume={357},
  number={22},
  pages={2277--2284},
  year={2007},
  publisher={Mass Medical Soc}
}

@article{nam2019development,
  title={Development and validation of deep learning--based automatic detection algorithm for malignant pulmonary nodules on chest radiographs},
  author={Nam, Ju Gang and Park, Sunggyun and Hwang, Eui Jin and Lee, Jong Hyuk and Jin, Kwang-Nam and Lim, Kun Young and Vu, Thienkai Huy and Sohn, Jae Ho and Hwang, Sangheum and Goo, Jin Mo and others},
  journal={Radiology},
  volume={290},
  number={1},
  pages={218--228},
  year={2019},
  publisher={Radiological Society of North America}
}

@ARTICLE{DeLong1988-zz,
  title    = "Comparing the areas under two or more correlated receiver operating characteristic curves: a nonparametric approach",
  author   = "DeLong, E R and DeLong, D M and Clarke-Pearson, D L",
  journal  = "Biometrics",
  volume   =  44,
  number   =  3,
  pages    = "837--845",
  month    =  sep,
  year     =  1988,
  address  = "England",
  language = "en"
}

@misc{chexagent,
      title={A Vision-Language Foundation Model to Enhance Efficiency of Chest X-ray Interpretation}, 
      author={Zhihong Chen and Maya Varma and Justin Xu and Magdalini Paschali and Dave Van Veen and Andrew Johnston and Alaa Youssef and Louis Blankemeier and Christian Bluethgen and Stephan Altmayer and Jeya Maria Jose Valanarasu and Mohamed Siddig Eltayeb Muneer and Eduardo Pontes Reis and Joseph Paul Cohen and Cameron Olsen and Tanishq Mathew Abraham and Emily B. Tsai and Christopher F. Beaulieu and Jenia Jitsev and Sergios Gatidis and Jean-Benoit Delbrouck and Akshay S. Chaudhari and Curtis P. Langlotz},
      year={2024},
      eprint={2401.12208},
      archivePrefix={arXiv},
      primaryClass={cs.CV},
      url={https://arxiv.org/abs/2401.12208}, 
}

@ARTICLE{mimic-cxr-paper,
  title    = "{MIMIC-CXR}, a de-identified publicly available database of chest
              radiographs with free-text reports",
  author   = "Johnson, Alistair E W and Pollard, Tom J and Berkowitz, Seth J
              and Greenbaum, Nathaniel R and Lungren, Matthew P and Deng,
              Chih-Ying and Mark, Roger G and Horng, Steven",
  journal  = "Scientific Data",
  volume   =  6,
  number   =  1,
  pages    = "317",
  month    =  dec,
  year     =  2019
}

@article{padchest,
   title={PadChest: A large chest x-ray image dataset with multi-label annotated reports},
   volume={66},
   ISSN={1361-8415},
   url={http://dx.doi.org/10.1016/j.media.2020.101797},
   DOI={10.1016/j.media.2020.101797},
   journal={Medical Image Analysis},
   publisher={Elsevier BV},
   author={Bustos, Aurelia and Pertusa, Antonio and Salinas, Jose-Maria and de la Iglesia-Vayá, Maria},
   year={2020},
   month=Dec, pages={101797} }

@misc{chexpert,
      title={CheXpert: A Large Chest Radiograph Dataset with Uncertainty Labels and Expert Comparison}, 
      author={Jeremy Irvin and Pranav Rajpurkar and Michael Ko and Yifan Yu and Silviana Ciurea-Ilcus and Chris Chute and Henrik Marklund and Behzad Haghgoo and Robyn Ball and Katie Shpanskaya and Jayne Seekins and David A. Mong and Safwan S. Halabi and Jesse K. Sandberg and Ricky Jones and David B. Larson and Curtis P. Langlotz and Bhavik N. Patel and Matthew P. Lungren and Andrew Y. Ng},
      year={2019},
      eprint={1901.07031},
      archivePrefix={arXiv},
      primaryClass={cs.CV},
      url={https://arxiv.org/abs/1901.07031}, 
}

@techreport{who_cancer_deaths,
  author      = {{International Agency for Research on Cancer}},
  title       = {World Fact Sheet: Global Cancer Observatory},
  institution = {World Health Organization},
  year        = {2024},
  type        = {Fact Sheet},
  url         = {https://gco.iarc.who.int/media/globocan/factsheets/populations/900-world-fact-sheet.pdf},
}

@ARTICLE{blind_spots,
  title    = "The blind spots on chest computed tomography: what do we miss",
  author   = "Zhang, Li and Wen, Xin and Ma, Jing-Wen and Wang, Jian-Wei and
              Huang, Yao and Wu, Ning and Li, Meng",
  journal  = "J Thorac Dis",
  volume   =  16,
  number   =  12,
  pages    = "8782--8795",
  year     =  2024,
  doi = {https://doi.org/10.21037/jtd-24-1125}
}

@ARTICLE{emphysema,
  title     = "Advances in imaging for lung emphysema",
  author    = "Martini, Katharina and Frauenfelder, Thomas",
  journal   = "Ann. Transl. Med.",
  publisher = "AME Publishing Company",
  volume    =  8,
  number    =  21,
  pages     = "1467",
  year      =  2020,
  doi={10.21037/atm.2020.04.44 }
}

@ARTICLE{design_cxr,
  title     = "Design and development of a new multi-projection X-ray system
               for chest imaging",
  author    = "Chawla, Amarpreet S and Boyce, Sarah and Washington, Lacey and
               McAdams, H Page and Samei, Ehsan",
  journal   = "IEEE Trans. Nucl. Sci.",
  publisher = "Institute of Electrical and Electronics Engineers (IEEE)",
  volume    =  56,
  number    =  1,
  pages     = "36--45",
  year      =  2009,
  copyright = "https://ieeexplore.ieee.org/Xplorehelp/downloads/license-information/IEEE.html",
}

@misc{mamba,
      title={Mamba: Linear-Time Sequence Modeling with Selective State Spaces}, 
      author={Albert Gu and Tri Dao},
      year={2024},
      eprint={2312.00752},
      archivePrefix={arXiv},
      primaryClass={cs.LG},
      url={https://arxiv.org/abs/2312.00752}, 
}

@misc{vmamba,
      title={VMamba: Visual State Space Model}, 
      author={Yue Liu and Yunjie Tian and Yuzhong Zhao and Hongtian Yu and Lingxi Xie and Yaowei Wang and Qixiang Ye and Jianbin Jiao and Yunfan Liu},
      year={2024},
      eprint={2401.10166},
      archivePrefix={arXiv},
      primaryClass={cs.CV},
      url={https://arxiv.org/abs/2401.10166}, 
}

@misc{chexnet,
      title={CheXNet: Radiologist-Level Pneumonia Detection on Chest X-Rays with Deep Learning}, 
      author={Pranav Rajpurkar and Jeremy Irvin and Kaylie Zhu and Brandon Yang and Hershel Mehta and Tony Duan and Daisy Ding and Aarti Bagul and Curtis Langlotz and Katie Shpanskaya and Matthew P. Lungren and Andrew Y. Ng},
      year={2017},
      eprint={1711.05225},
      archivePrefix={arXiv},
      primaryClass={cs.CV},
      url={https://arxiv.org/abs/1711.05225}, 
}

@misc{image_is_worth_16,
      title={An Image is Worth 16x16 Words: Transformers for Image Recognition at Scale}, 
      author={Alexey Dosovitskiy and Lucas Beyer and Alexander Kolesnikov and Dirk Weissenborn and Xiaohua Zhai and Thomas Unterthiner and Mostafa Dehghani and Matthias Minderer and Georg Heigold and Sylvain Gelly and Jakob Uszkoreit and Neil Houlsby},
      year={2021},
      eprint={2010.11929},
      archivePrefix={arXiv},
      primaryClass={cs.CV},
      url={https://arxiv.org/abs/2010.11929}, 
}

@misc{replace_cnn,
      title={Is it Time to Replace CNNs with Transformers for Medical Images?}, 
      author={Christos Matsoukas and Johan Fredin Haslum and Magnus Söderberg and Kevin Smith},
      year={2021},
      eprint={2108.09038},
      archivePrefix={arXiv},
      primaryClass={cs.CV},
      url={https://arxiv.org/abs/2108.09038}, 
}

@misc{dinocxr,
      title={DINO-CXR: A self supervised method based on vision transformer for chest X-ray classification}, 
      author={Mohammadreza Shakouri and Fatemeh Iranmanesh and Mahdi Eftekhari},
      year={2023},
      eprint={2308.00475},
      archivePrefix={arXiv},
      primaryClass={eess.IV},
      url={https://arxiv.org/abs/2308.00475}, 
}

@misc{umamba,
      title={U-Mamba: Enhancing Long-range Dependency for Biomedical Image Segmentation}, 
      author={Jun Ma and Feifei Li and Bo Wang},
      year={2024},
      eprint={2401.04722},
      archivePrefix={arXiv},
      primaryClass={eess.IV},
      url={https://arxiv.org/abs/2401.04722}, 
}

@misc{ihme_gbd_cancer_2025,
  author       = {{Institute for Health Metrics and Evaluation (IHME)}},
  title        = {GBD Cancer Compare Data Visualization},
  year         = {2025},
  howpublished = {IHME, University of Washington, Seattle, WA},
  url          = {https://vizhub.healthdata.org/gbd-compare/cancer},
  note         = {Accessed: 30 apr. 2026}
}

@Article{bias_1,
author={Seyyed-Kalantari, Laleh
and Zhang, Haoran
and McDermott, Matthew B. A.
and Chen, Irene Y.
and Ghassemi, Marzyeh},
title={Underdiagnosis bias of artificial intelligence algorithms applied to chest radiographs in under-served patient populations},
journal={Nature Medicine},
year={2021},
month={Dec},
day={01},
volume={27},
number={12},
pages={2176-2182},
issn={1546-170X},
doi={10.1038/s41591-021-01595-0},
url={https://doi.org/10.1038/s41591-021-01595-0}
}

@ARTICLE{scanner_bias,
  title     = "Variable generalization performance of a deep learning model to
               detect pneumonia in chest radiographs: A cross-sectional study",
  author    = "Zech, John R and Badgeley, Marcus A and Liu, Manway and Costa,
               Anthony B and Titano, Joseph J and Oermann, Eric Karl",
  journal   = "PLoS Med.",
  publisher = "Public Library of Science (PLoS)",
  volume    =  15,
  number    =  11,
  pages     = "e1002683",
  month     =  nov,
  year      =  2018,
  copyright = "http://creativecommons.org/licenses/by/4.0/",
  language  = "en"
}

@ARTICLE{Kelly2019-gr,
  title     = "Key challenges for delivering clinical impact with artificial
               intelligence",
  author    = "Kelly, Christopher J and Karthikesalingam, Alan and Suleyman,
               Mustafa and Corrado, Greg and King, Dominic",
  journal   = "BMC Med.",
  publisher = "Springer Science and Business Media LLC",
  volume    =  17,
  number    =  1,
  pages     = "195",
  month     =  oct,
  year      =  2019,
  copyright = "http://creativecommons.org/licenses/by/4.0/",
  language  = "en"
}

@inproceedings{pooch2020,
author = {Pooch, Eduardo H. P. and Ballester, Pedro and Barros, Rodrigo C.},
title = {Can We Trust Deep Learning Based Diagnosis? The Impact of Domain Shift in Chest Radiograph Classification},
year = {2020},
isbn = {978-3-030-62468-2},
publisher = {Springer-Verlag},
address = {Berlin, Heidelberg},
url = {https://doi.org/10.1007/978-3-030-62469-9_7},
doi = {10.1007/978-3-030-62469-9_7},
booktitle = {Thoracic Image Analysis: Second International Workshop, TIA 2020, Held in Conjunction with MICCAI 2020, Lima, Peru, October 8, 2020, Proceedings},
pages = {74–83},
numpages = {10},
location = {Lima, Peru}
}

\end{document}